%% file: jcsce-latex.tex
\documentclass[3p,times,twocolumn,12pt, procedia]{elsarticle}

\usepackage{ecrc}
\usepackage{float}
\usepackage[hang,flushmargin]{footmisc}
\usepackage{balance}

\volume{Vol. 38, No. 1}
\firstpage{11}
\journalname{VNU Journal of Science: Comp. Science $\&$ Com. Eng.,}

\runauth{Nghia et al.}

\usepackage[utf8]{inputenc}
\usepackage{babel}
\usepackage{mathtools}
\usepackage[T1]{fontenc}
\usepackage{graphicx}
\usepackage[labelsep=endash]{caption}
\usepackage{subcaption}
\usepackage{booktabs}
\usepackage{multirow}
\usepackage[none]{hyphenat}
\usepackage[tracking=true]{microtype}
\usepackage[labelsep=period,justification=centering]{caption}

\renewcommand\footnoterule{\kern-3pt \hrule width 0.65in \kern 2.6pt}
\usepackage{geometry}
\begin{document}
\begin{frontmatter}
\title{VieCap4H - VLSP 2021: ObjectAoA - Enhancing performance of Object Relation Transformer with Attention on Attention for Vietnamese image captioning}


\author{Nghia Hieu Nguyen$^{1,2}$}

\author{Duong T.D. Vo$^{1,2}$}

\author{Minh-Quan Ha$^{1,2}$}

\address{\normalsize $^{1}$University of Information Technology, Ho Chi Minh City, Vietnam\\
$^{2}$Vietnam National University, Ho Chi Minh City, Vietnam}

\cortext[label1]{\textls[-30]{~Corresponding author. Email.: 19520178@gm.uit.edu.vn}}

\begin{abstract}
\indent

Image captioning is currently a challenging task that requires the ability to both understand visual information and use human language to describe this visual information in the image. In this paper, we propose an efficient way to improve the image understanding ability of transformer-based method by extending Object Relation Transformer architecture with Attention on Attention mechanism. Experiments on the VieCap4H dataset show that our proposed method significantly outperforms its original structure on both the public test and private test of the Image Captioning shared task held by VLSP.
\end{abstract}

\end{frontmatter}

\input{sections/Introduction}
\input{sections/RelatedWorks}
\input{sections/TransformerBasedMethod}
\input{sections/OurProposedMethod}
\input{sections/Experiments}
\input{sections/Conclusion}

\balance
\bibliographystyle{elsarticle-num}

\bibliography{ref}

\appendix
\section{Example captions by ORT and ObjectAoA}
\input{sections/Example}

\end{document}

%% file: sections/Introduction.tex
\section{Introduction}
Image captioning is first introduced with the release of Flickr8k dataset \cite{hodosh2013framing}. This task has many practical applications: it can be used as a feature for automatically generating alternative text for images in social media applications; or can be used as a function in support devices for visually impaired people. Moreover, results from image captioning researches can be inherited to propose novel methods for visual question answering tasks. Regardless of its helpful application, image captioning is still challenging as it requires the model to understand visual features provided by the images and exploit the linguistic features of human language to describe the content of images.

Some early approaches tackled this task by using the combination of Convolutional Neural Network (CNN) to learn visual features from images and Recurrent Neural Network (RNN) to learn linguistic features from captions. With the advance of attention mechanisms, deep learning models can generate more accurate captions for images by adding attention features in their RNN structure \cite{xu2015show,vinyals2015show} or using Transformer structure \cite{vaswani2017attention} instead of RNN structure \cite{herdade2019image,he2020image,huang2019attention,cornia2020meshed}.

Although the improvement in linguistic aspect of captions generated from recently novel state-of-the-art models, these captions still have some drawbacks. Their meanings are somewhat general, in which a class of similar images can be given exactly the same caption. Moreover, considering more about captions, most of them describe the overall context in images, not go deeper into the details. These drawbacks draw a large challenge for image captioning models, requiring them to focus their attention on main objects in images and exploit more the human linguistic aspect to make better captions. To this end, Bottom-up Attention mechanism \cite{anderson2018bottom} is proposed to make deep learning model focus better on meaningful objects in the images, leaving the research community the opportunity to explore and propose the structure which can effectively use these attention-worthy visual features to learn how to make better captions.

Based on the work of \cite{anderson2018bottom}, many attention architectures were proposed to enhance the deep learning model to generate more meaningful captions \cite{huang2019attention,he2020image,cornia2020meshed}. These novel methods respectively achieved state-of-the-art (SOTA) results on the MS COCO dataset \cite{lin2014microsoft}. Although there are many mechanisms proposed in encoder layers of transformer-based methods to enhance image understanding ability, there is no work exploring how we can use these mechanisms together effectively. Hence in this work, we introduce a novel architecture which effectively uses together the two SOTA approaches proposed by \cite{huang2019attention} and \cite{herdade2019image}.

%% file: sections/RelatedWorks.tex
\section{Related works}
Vinyals et al. proposed the early approach, Show and Tell \cite{vinyals2015show} , for image captioning task by using CNN for learning features from images, then using RNN to learn how to make captions from learned visual features. Xu et al. improved the Show and Tell model \cite{xu2015show} by expanding its RNN structure with attention mechanisms \cite{bahdanau2014neural,luong2015effective}. After analysing visual aspect of captioning models, \cite{anderson2018bottom} proposed another attention stage at visual extraction module of image caption models, called Bottom-up attention instead of using grid features. Recently, Vaswani et al. \cite{vaswani2017attention} proposed Transformer architecture which have become the new research trend for the natural language processing (NLP) community. To take into account the advantage of Transformer architecture for linguistic aspect of captioning models, he et al. \cite{he2020image} proposed replacing RNN structure and its attention with transformer-based architecture. Huang et al. \cite{huang2019attention} discovered that the information returned from attention layer of transformer lack correlation with the information from the input, so they proposed another attention stage for Transformer architecture, called Attention on Attention (AoA). Herdade et al. \cite{herdade2019image} enhanced the attention module in transformer structure by proposing geometric attention mechanisms to make better understanding of visual features, so introduce a novel method called Object Relation Transformer. All the three transformer-based approaches, Image Transformer, AoA, and Object Relation Transformer, respectively achieved SOTA results on MS COCO dataset \cite{lin2014microsoft}.

For image captioning tasks in Vietnamese, there are quite limited works and experiments on Vietnamese. Recently, the first qualified image captioning dataset for Vietnamese UIT-ViIC \cite{lam2020uit} has been introduced to the research community. At the same time, \cite{10.1007/978-3-030-63119-2_64} published the first experiment for image captioning task in Vietnamese, but their approach was too trivial. They used machine translating models to translate caption from MS COCO dataset into Vietnamese then evaluated image captioning models on these captions, which have many drawbacks as analysed by \cite{lam2020uit}.

%% file: sections/TransformerBasedMethod.tex
\section{Transformer-based methods}
In this section, we briefly review the two SOTA transformer-based methods: Attention on Attention Network and Object Relation Transformer. Relied on the ideas of these methods, we then propose the novel approach (section 4) and analyse our proposed methods compared with its original architecture on both the UIT-ViIC dataset \cite{lam2020uit} and the vieCap4H dataset \cite{vlsp-2021-viecap4h} (section 5).

\subsection{Attention on Attention (AoA)}

\begin{figure}
    \centering
    \includegraphics[width=0.20\textwidth]{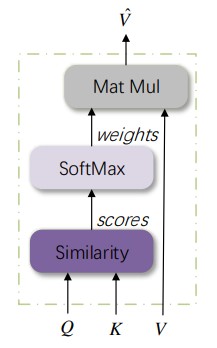}
    \includegraphics[width=0.20\textwidth]{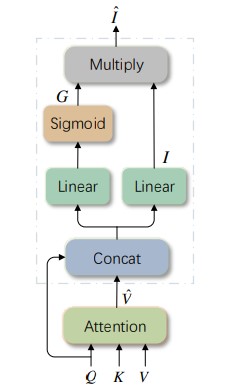}
    \caption{cross-attention (left) and AoA (right)}
    \label{fig:fig_1}
\end{figure}

Original cross-attention mechanism in transformer \cite{vaswani2017attention} (figure \ref{fig:fig_1}) forms the attention weight matrix as: 
$$ 
    \Omega_A = \displaystyle\frac{QK^T}{\sqrt{d_k}} 
$$
then the output of the attention layer is
$$
    f_{att}(Q, K, V) = Softmax(\Omega_A) V
$$
where $Q \in \mathbb{R}^{seq \times d_q}, K \in \mathbb{R}^{seq \times d_k}, V \in \mathbb{R}^{seq \times d_v}$ are the three input vectors of attention layer. Huang et al. \cite{huang2019attention} observed that the output weights can have low correlation with the input vectors as not all parts from captions have its correspondences in images. Therefore, we do not need to pay any attention at visual features for such these parts. To this end, they proposed to use additional linear layers as extended attention stage to ensure strong relation between the output of attention layer and the input vectors, reformulated the output of the attention layer as:
$$
    AoA(Q, K, V) = \sigma(G(Q, K, V)) \odot I(Q, K, V)
$$
where
$$
    G(Q, K, V) = W_q^g Q + W_v^g f_{att}(Q, K, V) + b^g
$$
$$
    I(Q, K, V) = W_q^i Q + W_v^i f_{att}(Q, K, V) + b^i
$$

\subsection{Object Relation Transformer (ORT)}

Inherited the advantage of Bottom-up attention \cite{anderson2018bottom}, \cite{herdade2019image} proposed to use additionally box coordinates achieved from RPN layer of FasterRCNN \cite{ren2015faster} to enhance the visual information from images. Given $Q \in \mathbb{R}^{seq \times d_q}, K \in \mathbb{R}^{seq \times d_k}, V \in \mathbb{R}^{seq \times d_v}$ the three input vectors of attention layer, the attention weights are calculated as:
$$
    \Omega_A = \displaystyle\frac{QK^T}{\sqrt{d_k}}
$$

The relative position of object $m$ to object $n$ in an image is represented as a 4-dimension vector: $$\lambda = \lambda(m, n) = (log(t_x), log(t_y), t_w, t_h)$$ where: $$ t_x = \displaystyle\frac{|x_m - x_n|}{w_m} $$ $$ t_y = \displaystyle\frac{|y_m - y_n|}{h_m} $$ $$ t_w = log\left(\frac{w_m}{w_n}\right) $$ $$ t_h = log\left(\frac{h_m}{h_n}\right) $$

Then the attention weight for the pair of object $m, n$ is given as: $$ w_{m, n} = ReLU(Emb(\lambda)W_G) $$

The final attention weight is calculated as: $$ \Omega = \Omega_A \circ exp(\Omega_W) $$ where $\Omega_W$ is the matrix formed by ${w_{m, n}}.$

%% file: sections/OurProposedMethod.tex
\section{Our proposal}

\begin{figure*}[ht]
    \centering
    \includegraphics[width=0.75\textwidth]{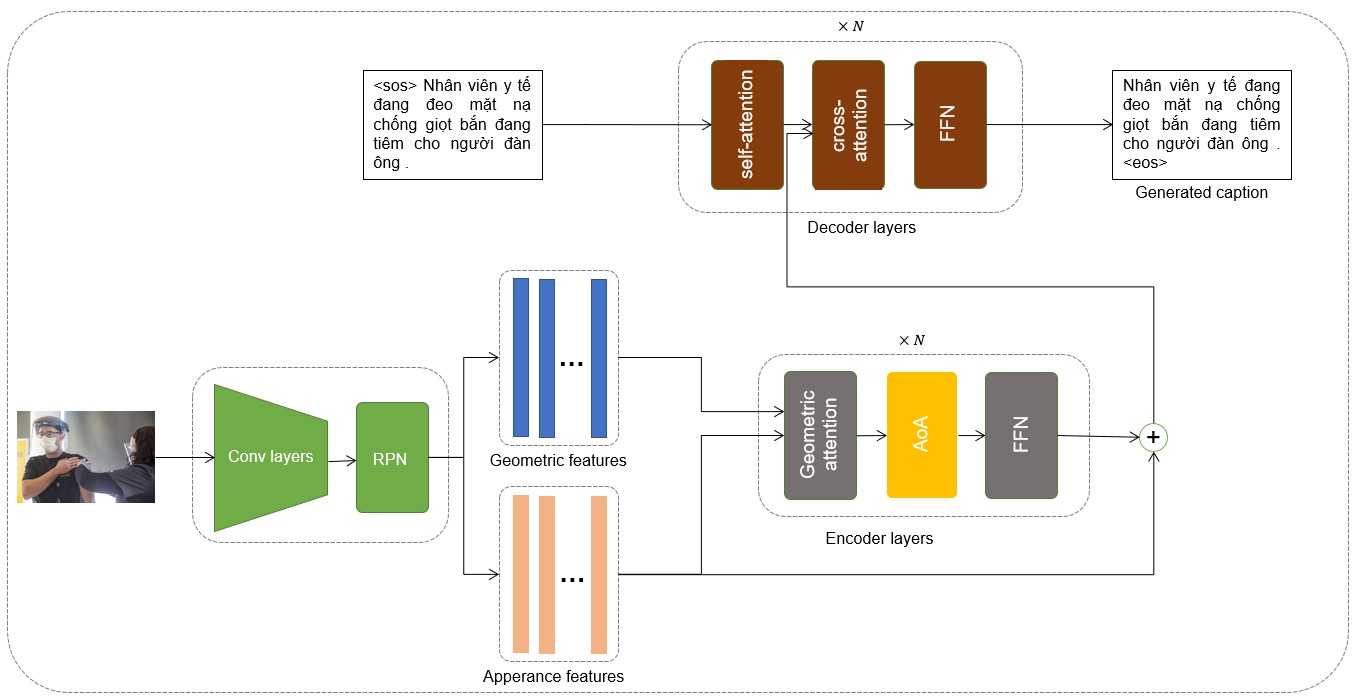}
    \caption{ObjectAoA Architecture (the yellow AoA module is the addition to the original ORT architecture)}
    \label{fig:fig_3}
\end{figure*}

Huang et al. \cite{huang2019attention} analysed the correlation between input information and attended information of self-attention module in transformer. But they equipped LSTM structure for their decoder layer which limited the ability to catch and understand linguistic features, potentially leading to lower performance. Herdade et al. \cite{herdade2019image} proposed to use geometric features of detected objects but they used the coordinates instead of area of bounding boxes. Moreover, the decoder of their model uses cross-attention mechanism to capture the human language rules, which can achieve better performance compared to the above methods. 

To this end, we propose to extend the utilization of geometric features with the idea of AoA to combine the advantages of these SOTA methods together. Specifically, from the attention weights $\Omega$ of geometric attention layer in Object Relation Transformer, the output of attention layer in our proposal is reformulated as:
$$
    \hat{V} = \sigma(G(Q, K, V)) \odot I(Q, K, V)
$$
where
$$
    G(Q, K, V) = W_q^g Q + W_v^g f_{att}(Q, K, V) + b^g
$$
$$
    I(Q, K, V) = W_q^i Q + W_v^i f_{att}(Q, K, V) + b^i
$$
$$
    f_{att}(Q, K, V) = \text{softmax}(\Omega) V
$$

As described above, our attention formulation is now equipped with an extra attention stage. We assume that the geometric attention exploits the geometric relationships between objects in images beside the internal relationships learned from visual feature vectors of these objects. However, not all the contexts in images contain the geometric relationship. Therefore, additional information about the locations of objects in images may somewhat cause confusion to the model when it learns how to use attention weights from the encoder layers in order to generate appropriate captions. This is when the Attention on Attention mechanisms come in handy. As analysed by \cite{huang2019attention}, Attention on Attention can increase the correlation between input information and output vectors from attention layers to create relevantly attended information. By applying Attention on Attention together with geometric attention, we can avoid uncorrelated output vectors of attention layers from the query vectors, making the model clearer when learning how to combine the information from visual and linguistic features, leading the model to generate better captions as a result. Moreover, to make the model better exploiting linguistic features, we use transformer as its decoder. For visual features, we inherit the advantage of region-based features yielded from Bottom-up Attention mechanisms \cite{anderson2018bottom} to let the model have the ability to focus on which objects it needs to describe. Finally, we named this method ObjectAoA \footnote{The implementation of ObjectAoA can be found in this repository
https://github.com/hieunghia-pat/UIT-ObjectAoA} (figure \ref{fig:fig_3}).

%% file: sections/Experiments.tex
\section{Experiments}
\subsection{Datasets}
In our experiments, we have evaluated the ORT method and our proposed method ObjectAoA on the UIT-ViIC dataset. Consequently, relying on the improvement on this dataset, we continue using the previous methods to train and evaluate on the VieCap4H dataset \cite{vlsp-2021-viecap4h} which was published in Image Captioning Shared Task by VLSP Challenge 2021. \footnote{Our results on the VieCap4H dataset can be found in the competition leaderboard at https://aihub.vn/competitions/68\#results}

\subsection{Metrics}
The popular metrics used for evaluating image captioning methods are BLEU \cite{papineni-etal-2002-bleu}, METEOR \cite{banerjee-lavie-2005-meteor}, ROUGE \cite{lin-2004-rouge}, CIDEr \cite{vedantam2015cider}. In our experiments, these four metrics are utilized to evaluate the two transformer-based models on the UIT-ViIC. For the VieCap4H datasets \cite{vlsp-2021-viecap4h}, we use the average of the four BLEU scores to evaluate ORT and our proposal.

\subsection{Training scheme}
We follow the work from Rennie et al. \cite{rennie2017self} where the training scheme is divided into two stages: in the first stage the model is trained using cross-entropy loss and in the second stage the model is refined using self-critical sequential training \cite{rennie2017self}. The self-critical sequential training scheme is a reinforcement learning strategy that consider the model as the "agent" and features from images and training captions are "signals" from the "environment". The rewards are calculated based on the CIDEr score in which the model will receive higher rewards if it can generate captions closer to the training captions.

\subsection{Configuration}
For first training training stage on the UIT-ViIC dataset, we trained the methods on the training set, used dev set to select the best models then evaluated them on the test set. All methods were trained on the GPU NVIDIA P100, using Adam \cite{kingma2014adam} optimizer with the learning rate adjusted follow the learning rate scheduler from Vaswani et al. \cite{vaswani2017attention}. The best models on dev set were chosen based on the CIDEr metric. We set the batch size to 32 and trained the models until the CIDEr score decreases.

For first training training stage on the VieCap4H dataset, we randomly split the provided training set into new training set having 7025 images and dev set having 1000 images. The best models were chosen based on average of the four BLEU metrics (BLEU-1, BLEU-2, BLEU-3 and BLEU-4). All methods were trained on the GPU NVIDIA P100, using Adam \cite{kingma2014adam} optimizer with the learning rate adjusted follow the learning rate scheduler from Vaswani et al. \cite{vaswani2017attention}. The best models on dev set were chosen based on the CIDEr metric. We also set the batch size to 32 and trained the models until the CIDEr score decreases.

For the self-critical training stage on both VieCap4H dataset and UIT-ViIC dataset, we fixed the learning rate to $5 \cdot 10^{-6}$ and trained the two models until the CIDEr score decreases.

\subsection{Experimental results}

\begin{table*}[ht]
    \centering
    \begin{tabular}{cccccccc}
        \hline
        \textbf{Methods} & \textbf{BLEU-1} & \textbf{BLEU-2} & \textbf{BLEU-3} & \textbf{BLEU-4} & \textbf{Meteor} & \textbf{Rouge} & \textbf{CIDEr} \\ \hline
        ORT              & 0.744           & 0.634  & 0.544            & 0.47           & 0.364           & 0.638          & 1.362          \\
        ObjectAoA        & \textbf{0.769}  & \textbf{0.66}           & \textbf{0.569}  & \textbf{0.497}  & \textbf{0.373}  & \textbf{0.648} & \textbf{1.488} \\ \hline
    \end{tabular}
    \caption{\label{table:1} Results of the transformer-based methods on the UIT-ViIC dataset.}
\end{table*}

We first use the UIT-ViIC dataset \cite{lam2020uit} to analyse the performance of our proposed method and ORT method. Results are showed in Table \ref{table:1}.

Results in Table \ref{table:1} indicate that ObjectAoA designed based on the Object Relation Transformer has better performance to its original architecture on all of the seven metrics. 

\begin{table*}[ht]
    \centering
    \begin{tabular}{cccc}
        \hline
        \textbf{Model}              & \textbf{dev} & \textbf{public test} & \textbf{private test} \\ \hline
        Object Relation Transformer & 0.3229      & 0.2645             & 0.2741                     \\
        ObjectAoA Transformer       & \textbf{0.3278}      & \textbf{0.295912}             & \textbf{0.285222}              \\ \hline
    \end{tabular}
    \caption{\label{table:2} Results of Object Relation Transformer and ObjectAoA Transformer on the VieCap4H dataset.}
\end{table*}

In our experiments on the UIT-ViIC dataset, we showed that our proposal yielded better results on the BLEU metric to other methods, therefore we used ObjectAoA as the main method to compete in the VLSP image captioning shared task. We concurrently conducted experiments for Object Relation Transformer on the VieCap4H dataset \cite{vlsp-2021-viecap4h} to compare with our proposal.

As Table \ref{table:2} indicates, the ObjectAoA Transformer outperforms the Object Relation Transformer in both dev set and test set. Relied on these results, we used ObjectAoA as our best method to submit to the private test phase.

The example captions generated by ORT and ObjectAoA are shown in figures \ref{fig:ex_12}, \ref{fig:ex_34}, \ref{fig:ex_56}, \ref{fig:ex_78} and \ref{fig:ex_910}. For each example image, there is a comparison between two corresponding captions given by two models. From the examples, it is clear that with the usage of AoA, the model can focus more precisely and describe the object correctly with better interpretation.

\subsection{Ablation study}
As depicted in figure \ref{fig:fig_3}, our proposed method ObjectAoA and the original ORT method are mostly the same, except for the AoA module in encoder layers. Therefore, the results on the UIT-ViIC dataset (table \ref{table:1}) and on the VieCap4H dataset (table \ref{table:2}) are the ablation study results for the effectiveness of the additional AoA module equipped at the encoder layers.

\subsection{Results analysis}
As depicted in Appendix A, most of the wrong cases of ORT method is catching wrong objects in the images (Fig.\ref{fig:ex_34} \#4, Fig.\ref{fig:ex_56} \#6, Fig.\ref{fig:ex_uit5}). Moreover, the ORT model failed to understand the context round objects in the images, which is a significant drawback causing ORT struggle to describe objects in images more specifically (Fig.\ref{fig:ex_12} \#1 and \#2, Fig.\ref{fig:ex_34} \#3, Fig.\ref{fig:ex_56} \#5, Fig.\ref{fig:ex_uit1}, Fig.\ref{fig:ex_uit2}, Fig.\ref{fig:ex_uit3}). We can realise this conclusion through the examples from the UIT-ViIC dataset, where captions generated from ORT model have general meaning compared to captions generated from ObjectAoA model which can describe both action and intention of objects in images. With the additional AoA module, the attention layer of ORT model is improved to enhance the correlation of the output features with the input features, which makes the ObjectAoA model can understand images better.

%% file: sections/Conclusion.tex
\section{Conclusion and future work}
In this paper, we have analysed and proposed a novel transformer architecture based on the two recent SOTA methods AoA \cite{huang2019attention} and ORT \cite{he2020image} for image captioning task. In the future, we will conduct more experiments and perform more deeper analysis on linguistic aspect of the ObjectAoA Transformer as well as extend other recent SOTA transformer-based models such as Meshed-Memory Transformer \cite{cornia2020meshed} with Attention on Attention mechanism to find out the better method for image captioning in Vietnamese. Moreover, we will develop the UIT-ViIC dataset to different domains and continue investigating image captioning architectures to finally conduct a cross-domain method for Vietnamese image captioning task.

%% file: sections/Example.tex
We note that as the captions of public test set and private test set of VieCap4H dataset \cite{vlsp-2021-viecap4h} are keep privately, we can only present the captions generated from ORT and ObjectAoA to compare the qualification of their results on this dataset.

\begin{figure*}[!ht]
    \centering
    \includegraphics[width=0.85\textwidth]{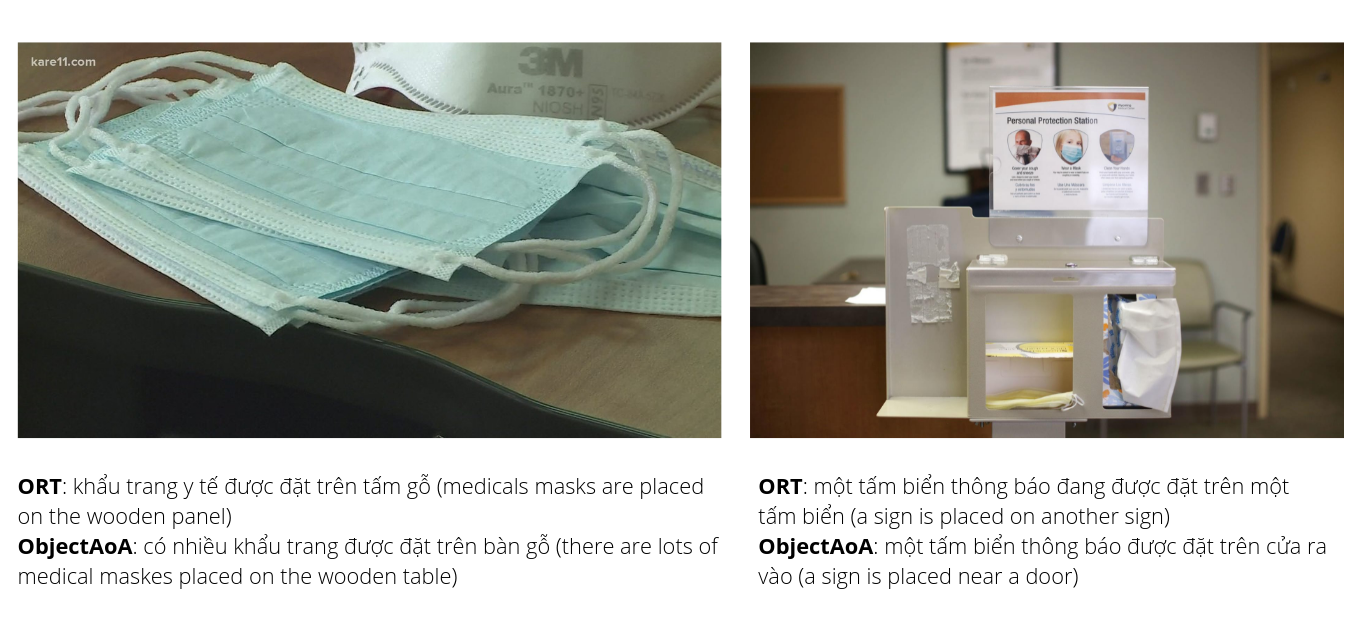}
    \caption{Example captions by ORT and ObjectAoA on the VieCap4H dataset \#1 and \#2}
    \label{fig:ex_12}
\end{figure*}

\begin{figure*}[!ht]
    \centering
    \includegraphics[width=0.85\textwidth]{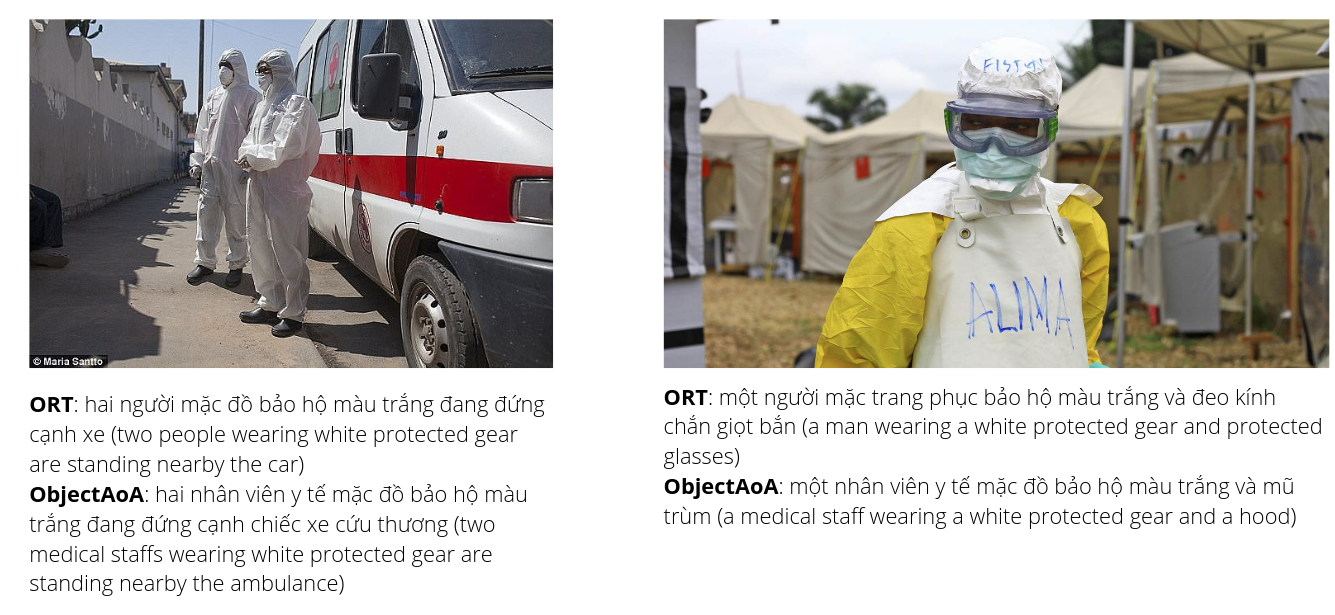}
    \caption{Example captions by ORT and ObjectAoA on the VieCap4H dataset \#3 and \#4}
    \label{fig:ex_34}
\end{figure*}

\begin{figure*}[!ht]
    \centering
    \includegraphics[width=0.85\textwidth]{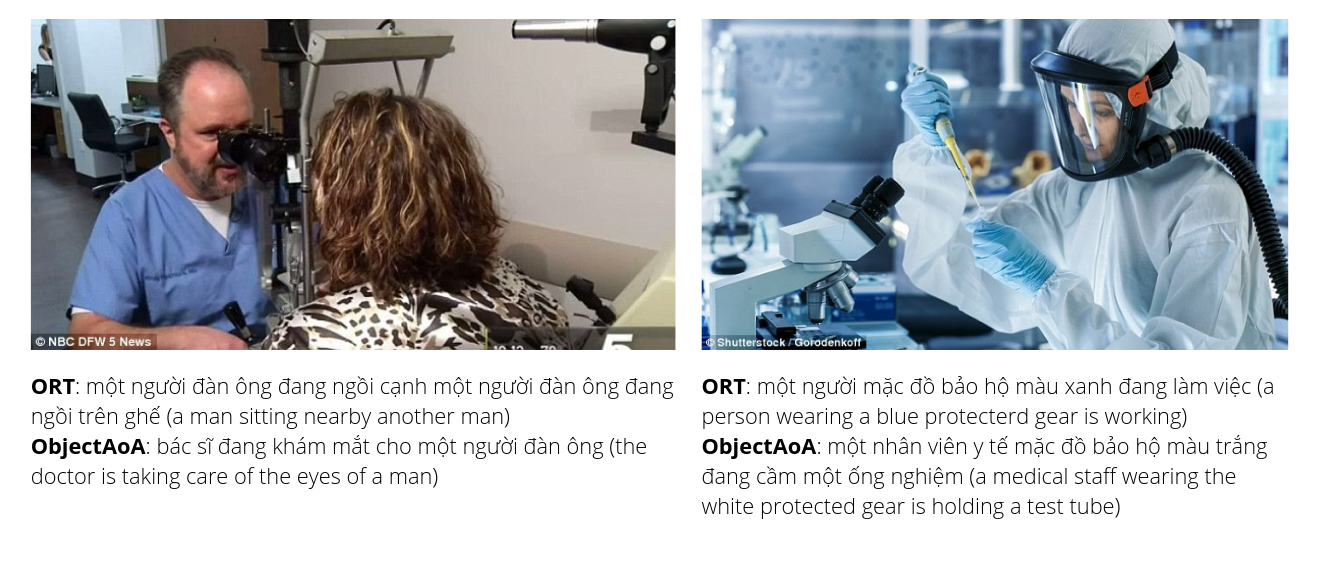}
    \caption{Example captions by ORT and ObjectAoA on the VieCap4H dataset \#5 and \#6}
    \label{fig:ex_56}
\end{figure*}

\begin{figure*}[!ht]
    \centering
    \includegraphics[width=0.85\textwidth]{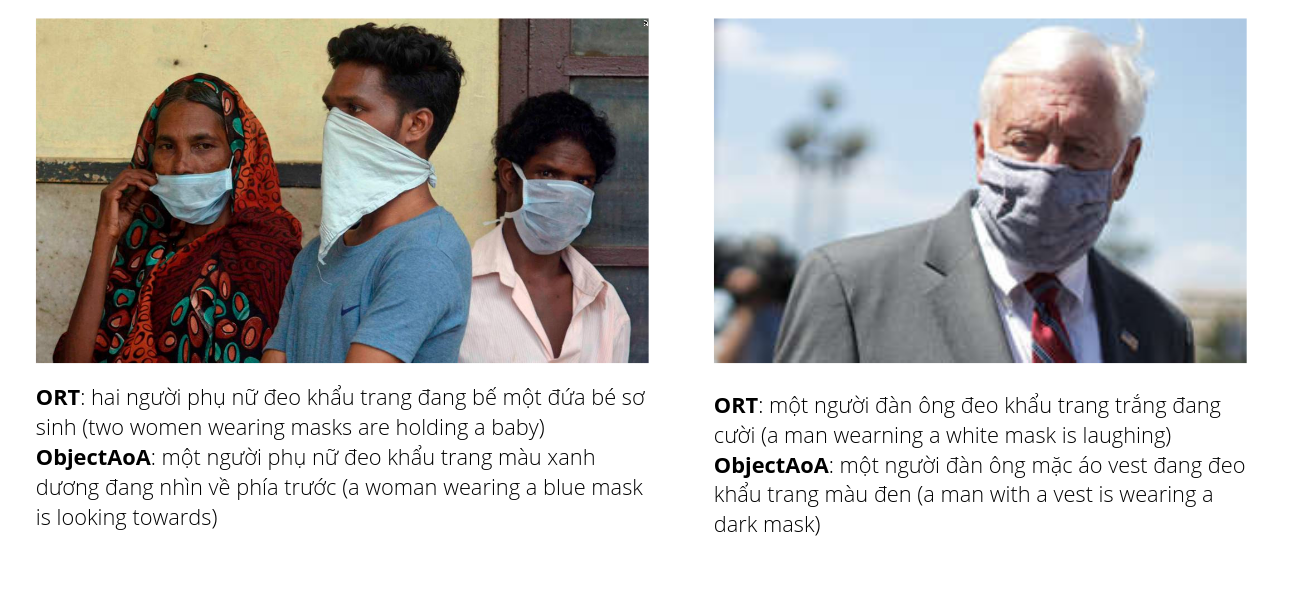}
    \caption{Example captions by ORT and ObjectAoA on the VieCap4H dataset \#7 and \#8}
    \label{fig:ex_78}
\end{figure*}

\begin{figure*}[!ht]
    \centering
    \includegraphics[width=0.85\textwidth]{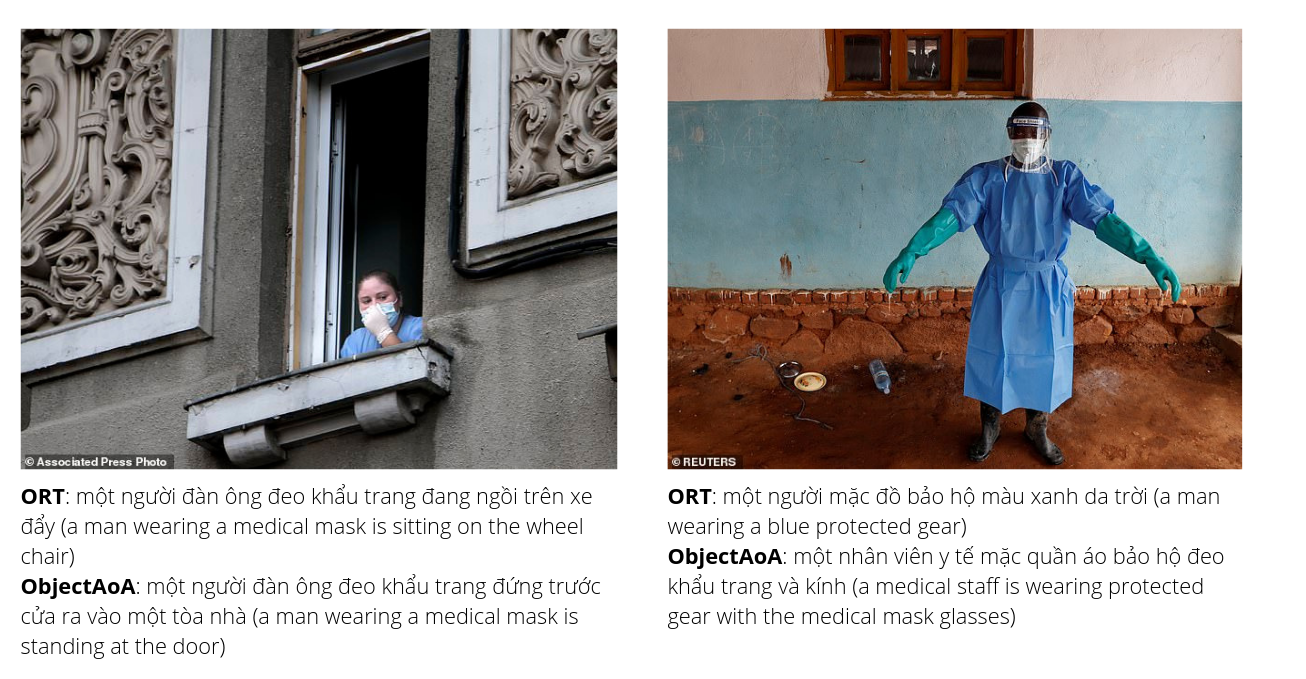}
    \caption{Example captions by ORT and ObjectAoA on the VieCap4H dataset \#9 and \#10}
    \label{fig:ex_910}
\end{figure*}

\begin{figure*}[!ht]
    \centering
    \includegraphics[width=0.85\textwidth]{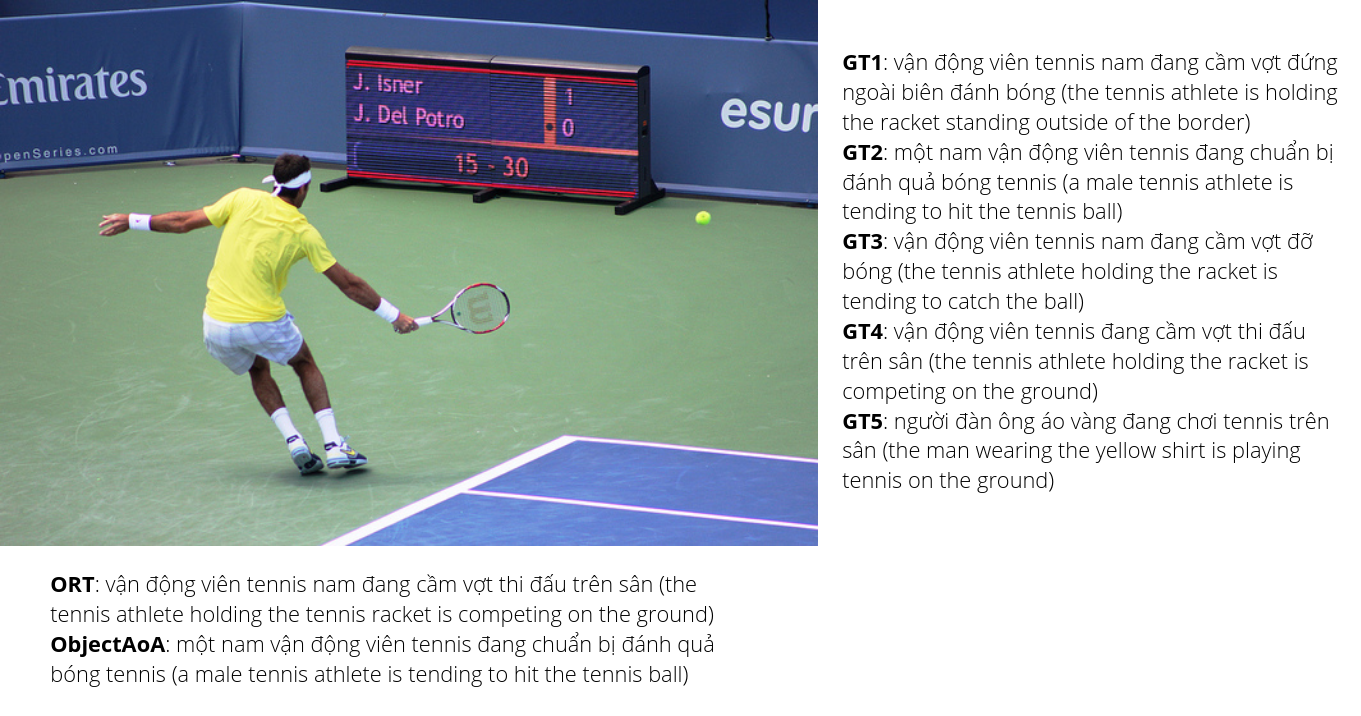}
    \caption{Example captions by ORT and ObjectAoA on the UIT-ViIC dataset \#1}
    \label{fig:ex_uit1}
\end{figure*}

\begin{figure*}[!ht]
    \centering
    \includegraphics[width=0.85\textwidth]{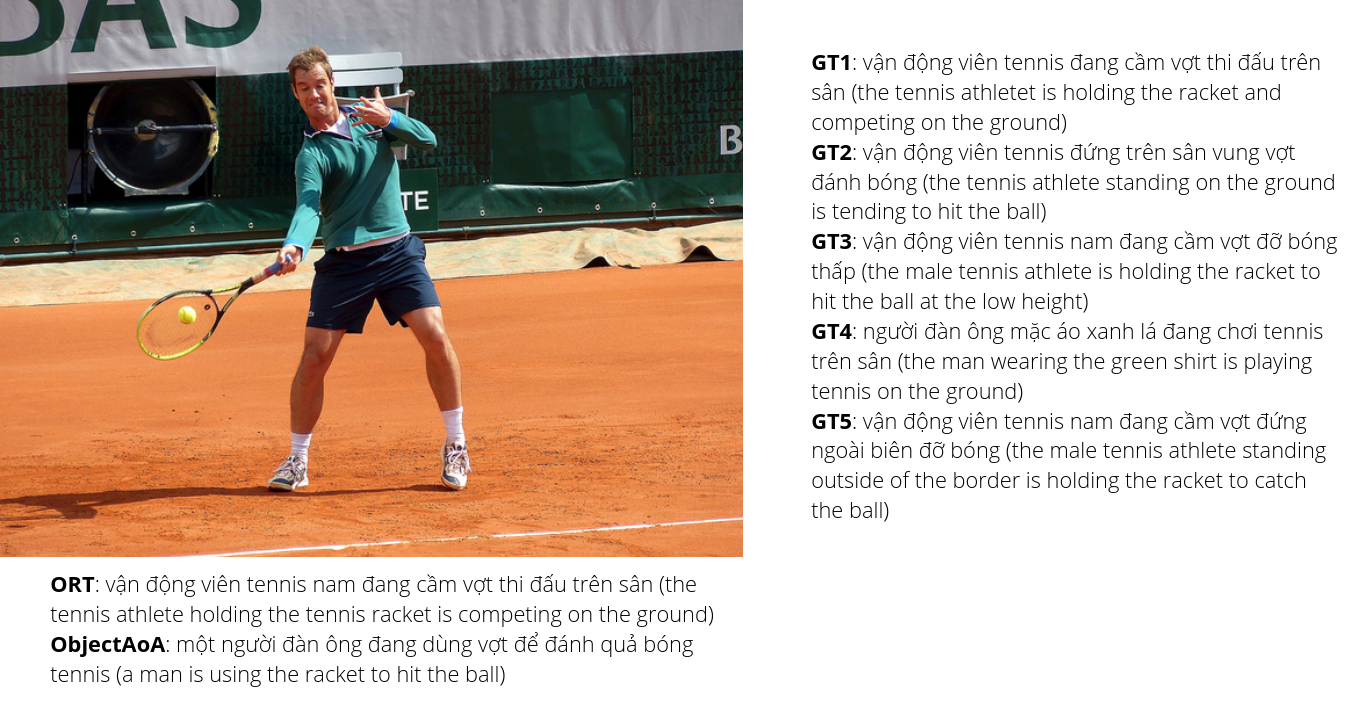}
    \caption{Example captions by ORT and ObjectAoA on the UIT-ViIC dataset \#2}
    \label{fig:ex_uit2}
\end{figure*}

\begin{figure*}[!ht]
    \centering
    \includegraphics[width=0.85\textwidth]{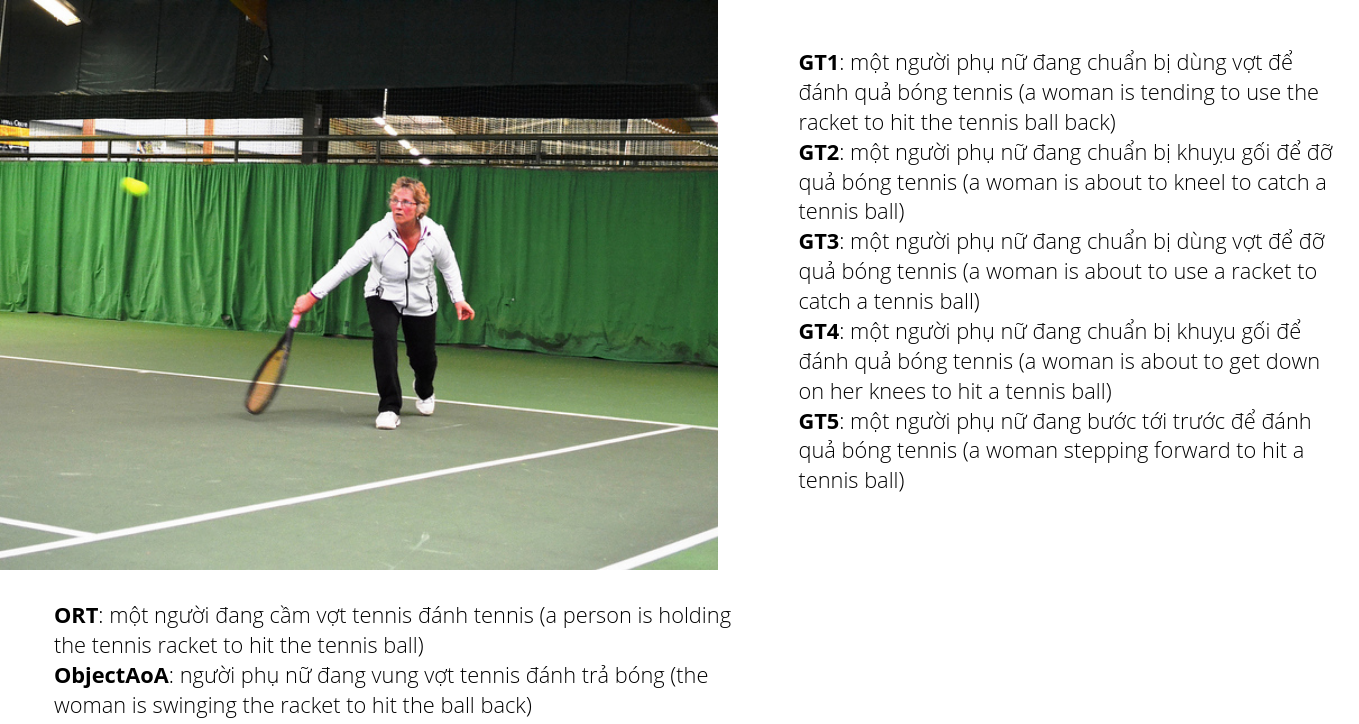}
    \caption{Example captions by ORT and ObjectAoA on the UIT-ViIC dataset \#3}
    \label{fig:ex_uit3}
\end{figure*}

\begin{figure*}[!ht]
    \centering
    \includegraphics[width=0.85\textwidth]{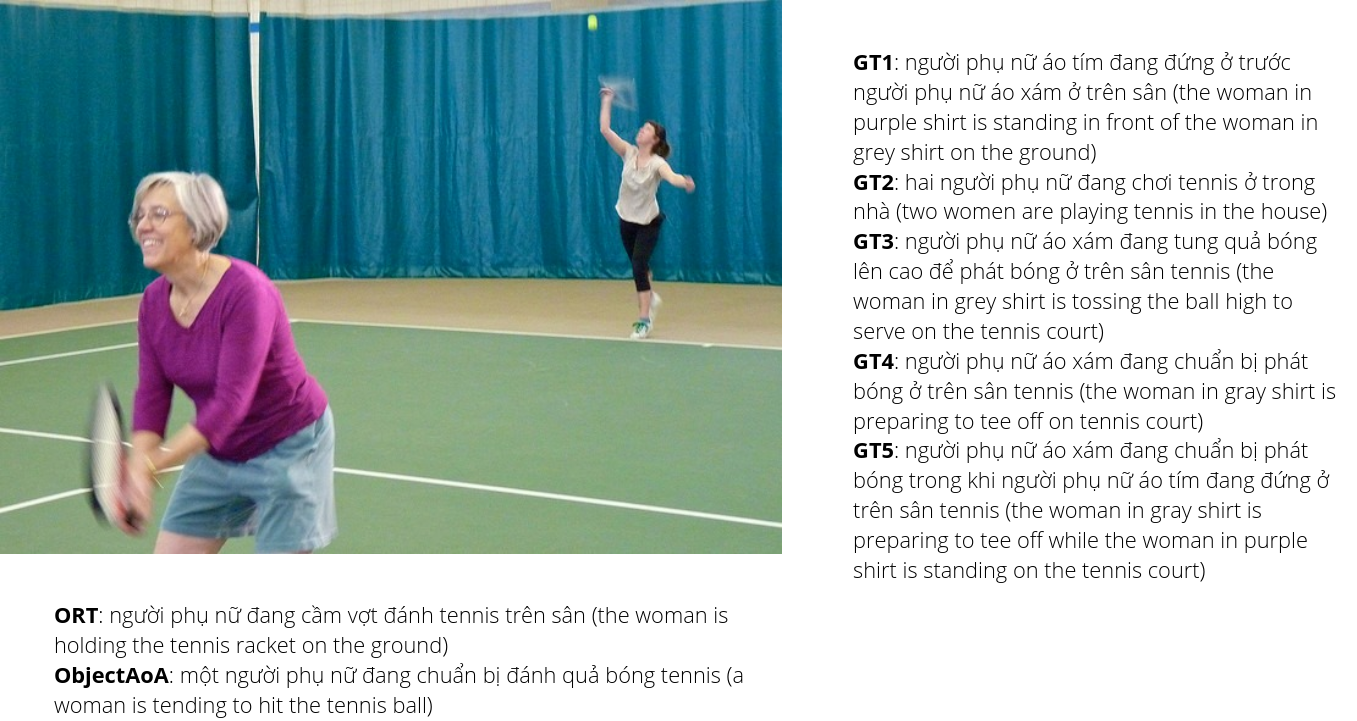}
    \caption{Example captions by ORT and ObjectAoA on the UIT-ViIC dataset \#4}
    \label{fig:ex_uit4}
\end{figure*}

\begin{figure*}[!ht]
    \centering
    \includegraphics[width=0.85\textwidth]{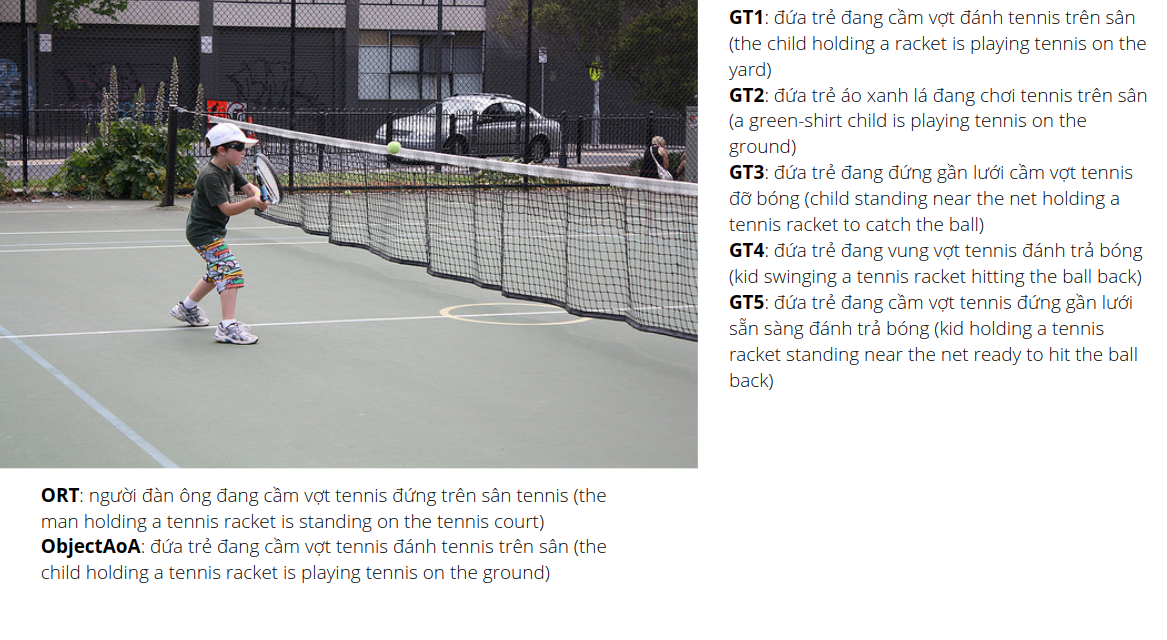}
    \caption{Example captions by ORT and ObjectAoA on the UIT-ViIC dataset \#5}
    \label{fig:ex_uit5}
\end{figure*}